\documentclass[conference]{IEEEtran}
\IEEEoverridecommandlockouts
\usepackage{cite}
\usepackage{amsmath,amssymb,amsfonts}
\usepackage{algorithmic}
\usepackage{graphicx}
\usepackage{textcomp}
\usepackage{xcolor}
\usepackage{url}

\usepackage{booktabs} 
\usepackage{array}    
\usepackage{tabularx}
\usepackage{array}
\newcolumntype{L}{>{\raggedright\arraybackslash}X}
\newcolumntype{C}{>{\centering\arraybackslash}X}
\newcolumntype{L}[1]{>{\raggedright\arraybackslash}m{#1}}
\newcolumntype{C}[1]{>{\centering\arraybackslash}m{#1}}

\usepackage[T1]{fontenc}
\usepackage{pifont}

\def\BibTeX{{\rm B\kern-.05em{\sc i\kern-.025em b}\kern-.08em
    T\kern-.1667em\lower.7ex\hbox{E}\kern-.125emX}}
\begin{document}

\title{InqEduAgent: Adaptive AI Learning Partners with Gaussian Process Augmentation\\

\thanks{This work was supported in part by Major Program of the National Natural Science Foundation of China under Grant 2026ZD1500300, in part by National Natural Science Foundation of China under Grant 62206112, and in part by Key Laboratory of Smart Education of Guangdong Higher Education Institutes under Grant 2022LSYS003.}
}

\author{
Wen-Xi Yang\textsuperscript{1}, 
Tian-Fang Zhao\textsuperscript{1,2}*, 
Guan Liu\textsuperscript{2}\\
\IEEEauthorblockA{\textsuperscript{1}\textit{Guangdong Institute of Smart Education}, \textit{Jinan University}, Guangzhou, China}

\IEEEauthorblockA{\textsuperscript{2}\textit{School of Journalism and Communication}, \textit{Jinan University}, Guangzhou, China}

\IEEEauthorblockA{*Corresponding author: tfzhao@jnu.edu.cn}
}


\maketitle

\begin{abstract}
Collaborative partnerships play a crucial role in inquiry-oriented education. However, most learning partners are currently assigned through experience-driven heuristics or rule-based machine assistants, which often result in limited knowledge expansion and low adaptability. To address these challenges, this study introduces InqEduAgent, an LLM-empowered generative agent framework designed to simulate and select adaptive learning partners for inquiry-based learning. InqEduAgent integrates a Gaussian process–augmented matching mechanism to model the cognitive and evaluative characteristics of learners, allowing adaptive partner selection based on prior knowledge patterns. Comprehensive experiments demonstrate that InqEduAgent consistently achieves superior performance across diverse learning scenarios and large language model configurations. This study advances human–AI collaborative learning by enabling intelligent pairing between human- and AI-based learning partners, and contributes to adaptive user modeling and personalized recommendation within Web-based educational environments. 
\end{abstract}


\section{Introduction}
Inquiry-oriented education has gained significant traction as an effective approach for fostering metacognitive skills, including critical thinking, self-regulated learning, questioning, and explanation \cite{azevedo2009metatutor,aleven2002effective}
Collaborative partnerships are a typical form of human–AI collaboration and play a crucial role in this paradigm.

Existing human-AI collaborations can be classified into three major categories. The first category includes upgraded versions of classic human–computer interaction systems, including personalized recommendation systems \cite{xia2020design}, AI-assisted home tutoring services \cite{kim2022home}, computer-supported cooperative work  \cite{wang2020human}, and AI Assistance in Home Tutoring Services. These platforms are capable of generating personalized learning programs and plans, yet they regard machines as mere objects while overlooking the two-way interaction between learners and machines. 

AI tutoring systems are the second category of research.
Examples include medical surgical simulation technology instructors \cite{fazlollahi2022effect}, programming tutoring systems \cite{ilic2024predicting}, AI instructors for older adults \cite{aly2024perceived}, and AI instructors’ communication styles \cite{kim2021like}.
These systems, seeing AI as a supertool, aim to enhance the teaching and learning experience in AI-enabled educational contexts. However, they rely heavily on predefined models, algorithms or  the accuracy of the data used for training, which are hard to adapt to unforeseen learning situations and student needs that deviate from the norm.

Human–AI co-learning patterns represent the third category and a future research direction. Studies in this direction focus on cooperative learning strategies, comparisons between AI and human teammates, and trustworthy human–AI partnerships\cite{zhang2023trust,ramchurn2021trustworthy,magnisalis2011adaptive}. These studies show that dynamic shifts between individual and collaborative learning can lead to a more comprehensive educational experience. This study falls into the third category of research. However, two key challenges remain in this area. Designing personalized AI co-learners that can enhance learners’ academic performance in terms of personality, preference, and learning efficiency remains a formidable problem. 
In addition, there is an ongoing and long-term ethical debate regarding trust in AI teammates and the extent to which AI co-learners should replace human learning partners.

Large language models (LLMs) have become a transformative force in intelligent education systems. A key milestone in generative agent research occurred in 2023 when Jaehyuk Park et al. \cite{park2023generative} introduced a Sims-like sandbox environment called Smallville (also known as Stanford Town) and created 25 generative agents. Since then, many studies have applied this idea to different domains, including simulation of human behavior \cite{park2023generative} and smart hospitals \cite{li2024agent}. Recent studies highlight the potential of generative agents in smart education \cite{baidoo2023education}. In 2023, Liu et al. proposed a thought-provoking teaching paradigm called SocraticLM \cite{liu2024socraticlm}. The chain-of-thought prompting approach for LLMs \cite{cohn2024chain} has also been applied to evaluate the formsative assessment responses of learners in science education. Then, task-specific generative agents have been developed, such as Agent4Edu \cite{gao2025agent4edu}. As LLMs have significantly improved their capabilities in autonomous interaction and decision-making, demonstrating a strong potential to simulate human social behaviors and serve as intelligent learning partners \cite{park2023generative}. It is necessary to re-examine how online learning systems for emotional interaction and social collaboration can flexibly coexist and transition between each other. Clarifying the functional boundaries and collaborative roles of these systems may help alleviate trust concerns about AI learning partners. With these considerations, this study contributes in two aspects.

1) Constructing a generative agent model, InqEduAgent, driven by semantic understanding and nonparametric modeling to simulate study partners. Agents are endowed with personas that match the cognitive characteristics of real-world learners. A mirror mapping strategy is introduced. Nonparametric modeling is used to map the textual features of natural language to numerical features. Prompt engineering is applied to map the numerical features back into text-based features, which drive the decision making of the agent.
Compared with traditional parameter-based models and purely semantic-oriented large language models, this model achieves effective parameterization and inverse parameterization.

2) Enhancing the Gaussian process with domain knowledge and the Pareto front to address feature pattern mining and domain matching. The learner characteristics and the exercise characteristics are mapped into a high-dimensional feature space. The kernel function is used to compute the covariance structure of the feature vectors. This process quantifies the matching distribution between different study partners and the target learner, forming a probability-based feature similarity assessment. The Pareto front is then introduced to preliminarily screen an optimal candidate set based on nondominated score vectors. The candidate with the highest predicted value is selected as the learning partner. Experimental results show that the proposed model performs well in multiple scenarios and effectively guides the agent in achieving optimal learning-partner matching.

\section{Proposed Model}
\label{sec:1}
\begin{figure*}[t]
\centering
\includegraphics[width=\linewidth]{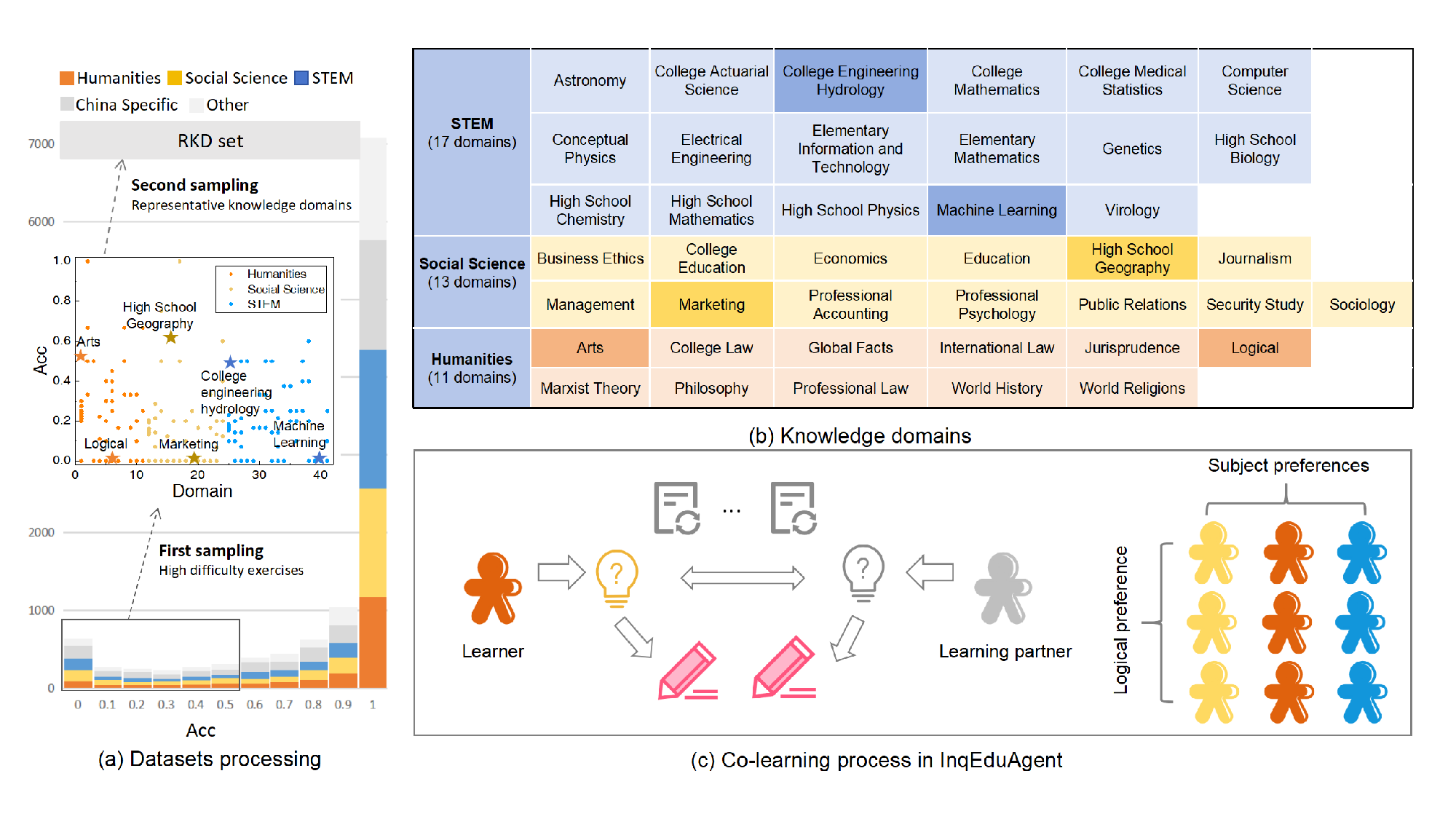}  
\caption{ The overall framework of InqEduAgent}
\label{fig:best_agent}
\end{figure*}

In this section, InqEduAgent is introduced as an inquiry-oriented education agent with Gaussian process enhancement.
As shown in Fig.~\ref{fig:best_agent}, InqEduAgent contains two primary components.
The first component applies nonparametric modeling to extract content structures from limited paired answering records and uncover learning and pairing patterns.
The second component integrates a Pareto set to constrain the search space and improve the orientation of the learning goals.

\subsection{Agentetic  Characters}
\label{sec:2}

A generative agent model is developed to simulate students in an intelligent education classroom.
To identify the core characteristics of agents and eliminate interfering factors, each agent is considered using only two characteristic dimensions: subject preference and logical preference.

Agents instantiated with assigned roles are referred to as learners.
Let $s \in \{-1,0,1\}$ denote subject preference and $g \in \{-1,0,1\}$ denote reasoning-logic preference.
A learner is parameterized as $\mathbf{A}(s,g)$, where $s=-1$ indicates a bias toward Social Sciences and Humanities,
$s=1$ indicates a bias toward STEM, and $s=0$ indicates no salient subject preference; $g=-1$ corresponds to deductive reasoning,
$g=1$ corresponds to inductive reasoning, and $g=0$ corresponds to intuitive reasoning with no salient preference.
The learner set is defined as
\begin{equation}
\begin{split}
L \triangleq \left\{\, \mathbf{A}(s,g) \ \middle|\ s\in\{-1,0,1\},\ g\in\{-1,0,1\} \right\}.
\end{split}
\end{equation}
\subsection{Environmental Interaction}
\label{sec:2}

In the intelligent education classroom, environmental interaction is defined as the interaction between learners and questions, as well as the interaction among learners.
In this paper, interactions between learners and exercises mean that, for each single-choice exercise, every learner independently thinks according to their own subject preference and logical preference.
Each learner then chooses an option and gives an answer along with a response text.
Interactions between learners refer to the situation in which two learners communicate with each other about their thoughts on the question itself, excluding the options, and exchange opinions.
Subsequently, they summarize based on their own cognitive characteristics and answer the question again.

\textbf{Learner–Exercise Interaction.}
For each learner $l \in L$, the interaction with an exercise $e \in E$ is modeled as a two-step process.
First, the learner independently reasons about the question $e$ together with its options based on its own cognitive characteristics.
Second, the learner outputs a final answer.
Each exercise $e$ is represented by an $M$-dimensional feature vector that encodes the associated knowledge concepts, denoted as
$\mathbf{e} = (e^1, e^2, \cdots, e^M), \quad m = 1,2,\ldots,M.$
The response of learner $l$ to exercise $e$ is recorded as a triplet
\begin{equation}
\mathbf{R}_{l}(e) = (l, \mathbf{e}, r_{l,e}),
\end{equation}
where $r_{l,e} \in \{0,1\}$ indicates whether learner $l$ answers exercise $e$ correctly.

\textbf{Learner–Learner Interaction.}
For any pair of learners $(l_i, l_j)$, their interaction is divided into four steps.
First, each learner independently thinks about the exercise $e'$, which contains only the question and no answer options.
Second, the two learners exchange their response texts as their thinking content.
Third, each learner summarizes the information exchanged based on their own cognitive characteristics to generate a summary text.
Fourth, each learner rethinks the exercise $e$ with answer options by considering the summary text and then answers the question again.
The response data at this stage are described as
\begin{equation}
\mathbf{R}_{l,\ell}(e) = ((l, \ell, e'), e, r_{l,\ell}),
\end{equation}
where $l \in L$, $\ell \in L \setminus \{l\}$, $e \in E$, and $r_{l,\ell} \in \{0,1\}$.
The difference between $e$ and $e'$ is that $e'$ removes the answer options to minimize the prior bias during communication.
This design helps prevent the transmission of correct or incorrect options that can interfere with the accuracy of the final answer.
In this process, learners communicate by exchanging their interpretations based on subject preferences and logical reasoning.
This form of interaction is called inquiry-based communication.
\subsection{ Gaussian Process Augmentation}
\label{sec:2}
A Gaussian process is a nonparametric statistical modeling method that is commonly used to model relationships between data points and to predict output values.
Based on empirical observations, the characteristics of learners and exercises follow approximately normal distributions.
These properties give the Gaussian process favorable mathematical characteristics.
As a result, it is well suited for modeling uncertainty and complex functions.

\textbf{Prior Knowledge and Embedding.}
To model the learning effects arising from learner interactions, it is necessary to transform the learner characteristics and exercise information into structured input samples suitable for Gaussian process modeling. In this work, a pre-trained BERT model is employed to obtain semantic representations of exercises, while learner characteristics are encoded into low-dimensional feature vectors.
Specifically, learner features are first mapped into a compact vector space reflecting their cognitive characteristics. Exercise texts are then embedded into a high-dimensional semantic space using BERT. The concatenation of learner and exercise embeddings serves as the input representation for subsequent modeling.
Based on domain knowledge, we design a Gaussian process to capture learning gains emerging from learner interactions. Rather than modeling individual questions independently, exercises belonging to the same knowledge domain are grouped into a domain-specific question set, denoted as $d$, which represents a coherent knowledge block.
Each training sample is represented as $x = (l, \ell, d)$, where $l$ denotes the target learner, $\ell$ denotes the interacting learner, and $d$ represents a set of exercises belonging to the same knowledge domain.
The corresponding output value measures the average learning gain obtained through interaction and is defined as
\begin{equation}
\Delta_{l,\ell,d}= \frac{1}{|d|} \biggl( 
\sum_{e \in d} r_{l,\ell,e} - \sum_{e \in d} r_{l,e} 
\biggr),
\end{equation}
where $r_{l,\ell,e}$ denotes the response score of learner $l$ on exercise $e$ after interacting with learner $\ell$, and $r_{l,e}$ denotes the score obtained by learner $l$ when solving the same exercise independently. Here, $d = \{ e_i \mid i = 1, \ldots, |d| \}$ 
represents the set of exercises within the domain. The value $y$ thus quantifies the average performance improvement attributable to learner interaction within a given knowledge domain.

\textbf{Non-parametric modeling.}
In collaborative learning scenarios, this study focuses on modeling the learning gain induced by communication rather than the absolute performance of learners.
Specifically, the goal is to estimate how much improvement a learner can achieve by interacting with different communication partners in a given learning context.

\textbf{Problem Formulation.}
Given a target learner $l$, a communication partner $\ell$, and a knowledge domain $d$, each interaction is modeled as a supervised sample.
The input vector is defined as
$\mathbf{x}_{l,\ell,d} = (\mathbf{A}_l, \mathbf{A}_\ell, \mathbf{e}_d, b_{l,d})$, where $\mathbf{A}_l$ and $\mathbf{A}_\ell$ denote the individual characteristics of learner $l$ and partner $\ell$, respectively.
The term $\mathbf{e}_d$ represents the semantic embedding of the question set in domain $d$, and $b_{l,d}$ denotes the baseline performance of learner $l$ in this domain before communication. The corresponding output variable is defined as the learning gain
$\Delta_{l,\ell,d} = p_{l,d}^{(\ell)} - b_{l,d}$
where $p_{l,d}^{(\ell)}$ denotes the performance achieved by learner $l$ in domain $d$ after interacting with partner $\ell$.
This definition directly captures the improvement attributable to the communication process.
Let $\mathcal{D} = \{(\mathbf{x}_i, \Delta_i)\}_{i=1}^{n}$ denote the set of observed interaction samples.

\textbf{Gaussian Process Modeling.}
The learning gain is assumed to be generated by an underlying latent function corrupted by independent Gaussian noise:
\begin{equation}
\Delta_i = f(\mathbf{x}_i) + \varepsilon_i, \quad \varepsilon_i \sim \mathcal{N}(0, \sigma_n^2),
\end{equation}
where $f(\cdot)$ models the latent relationship between learner–partner–task configurations and the induced learning gain.
The latent function is modeled as a Gaussian process:
\begin{equation}
f(\mathbf{x}) \sim \mathcal{GP}\big(m(\mathbf{x}), k(\mathbf{x}, \mathbf{x}')\big),
\end{equation}
where $m(\mathbf{x})$ is the mean function and $k(\mathbf{x}, \mathbf{x}')$ is the covariance function.
In this work, the radial basis function (RBF) kernel is adopted to capture smooth variations in learning gain across different interaction settings.
All observed samples are stacked into vector form as
$\boldsymbol{\Delta} = \mathbf{f} + \boldsymbol{\varepsilon}$,
where $\mathbf{f} = [f(\mathbf{x}_1), \ldots, f(\mathbf{x}_n)]^T$ and $\boldsymbol{\varepsilon} \sim \mathcal{N}(\mathbf{0}, \sigma_n^2 \mathbf{I})$.
Under the Gaussian process assumption, the joint distribution is given by
$\mathbf{f} \sim \mathcal{N}(\mathbf{m}, \mathbf{K})$,
where $\mathbf{m}$ is the mean vector and $\mathbf{K}$ is the covariance matrix with entries
$K_{ij} = k(\mathbf{x}_i, \mathbf{x}_j) + \sigma_n^2 \delta_{ij}$.
Model parameters are learned by maximizing the log marginal likelihood with respect to the kernel hyperparameters and the noise variance.

\textbf{Learning Gain Prediction.}
For a new candidate interaction configuration $\mathbf{x}^*$, the predictive distribution of the learning gain is modeled as a Gaussian distribution,
$p(\Delta^* \mid \mathbf{x}^*, \mathcal{D}) \sim \mathcal{N}(\mu^*, \sigma^{*2})$.
The predictive mean and variance are given by
$\mu^* = m(\mathbf{x}^*) + \mathbf{k}^{*T}\mathbf{K}^{-1}(\boldsymbol{\Delta} - \mathbf{m})$
and
$\sigma^{*2} = k(\mathbf{x}^*, \mathbf{x}^*) - \mathbf{k}^{*T}\mathbf{K}^{-1}\mathbf{k}^* + \sigma_n^2$,
respectively.
Here, $\mathbf{k}^* = [k(\mathbf{x}^*, \mathbf{x}_1), \ldots, k(\mathbf{x}^*, \mathbf{x}_n)]^T$ denotes the covariance vector between the new input $\mathbf{x}^*$ and the observed interaction samples.
The predictive mean $\mu^*$ is interpreted as the expected learning gain under the given interaction configuration and is used as the primary criterion for communication partner selection.

\textbf{Pareto-based Decision Strategy. } 
The primary objective of communication partner selection is to improve learners’ response accuracy across multiple knowledge domains.
To achieve this goal, a Pareto-based decision strategy is introduced to identify an optimal candidate set before final partner selection.
Specifically, the performance of a learner $l$ across multiple knowledge domains is represented by a score vector
$\mathbf{S}_l = (s_{l1}, s_{l2}, \dots, s_{lm})$,
where $s_{lk}$ denotes the score of learner $l$ in the $k$-th knowledge domain.
Similarly, the performance vector of a candidate communication partner $\ell$ is defined as
$\mathbf{S}_\ell = (s_{\ell1}, s_{\ell2}, \dots, s_{\ell m})$.
The score vector $\mathbf{S}_l$ is said to dominate $\mathbf{S}_\ell$ if learner $l$ performs no worse than $\ell$ in all knowledge domains and strictly better in at least one domain.
In this case, $\mathbf{S}_\ell$ is considered inferior.
All score vectors that are not dominated by any other candidate form the Pareto front.
This set serves as the optimal candidate pool for communication partner selection.
Instead of selecting from all possible partners, the final communication partner is chosen from the Pareto front by maximizing the predicted learning gain estimated by the Gaussian process model:
\begin{equation}
\ell^* = \operatorname{arg\,max}_{x \in \mathrm{ParetoFront}} \mu(x).
\end{equation}

In this paper, two Pareto-based criteria are adopted for partner selection: global Pareto and local Pareto.
Under the global Pareto criterion, the score vectors of all learners across multiple knowledge domains are treated as the candidate set.
Non-dominated score vectors in this set form the global Pareto front.
In the InqEduAgent model with global Pareto (InqEduAgent-GP), all learners select communication partners from this shared global Pareto front.
In contrast, the local Pareto criterion constructs a personalized candidate set for each learner.
During the training process of InqEduAgent, a learner interacts with multiple communication partners, and the resulting score vectors are collected as the candidate set.
Non-dominated score vectors are then selected to form a local Pareto front for that learner.
In the InqEduAgent model with local Pareto (InqEduAgent-LP), each learner maintains an individual Pareto front for partner selection.

\section{Experiments}
\subsection{Dataset analysis}
To evaluate the adaptability and effectiveness of InqEduAgent in a multidisciplinary knowledge setting, this study uses the CMMLU dataset \cite{li2024cmmlu} for experimentation.
CMMLU is a comprehensive evaluation benchmark that covers 67 knowledge domains across five major categories: STEM, social sciences, humanities, China-specific studies, and other fields.
It provides a challenging multi-task, multi-domain testing platform and is widely used to assess the knowledge coverage and reasoning capabilities of large language models.
A total of 11,582 exercises were evaluated using the Qwen-32B and DeepSeek-32B models.
For each exercise, the accuracy score was obtained by averaging the results over 10 repeated runs across the two models.

Fig.~\ref{fig:best_agent}(a) illustrates the distribution of question difficulty across all knowledge domains.
The number of exercises in the five major categories is roughly balanced, with STEM showing a slight numerical advantage.
The difficulty distribution of the exercises appears to follow an exponential pattern.
LLMs achieve a perfect score of 1 on approximately 61.1\% of the exercises and reach the passing threshold on about 82.8\% of the exercises.
These results suggest that existing LLMs have effectively learned and internalized a substantial amount of knowledge and problem-solving strategies relevant to the exercise content.

LLMs fail to pass 17.2\% of the exercises in the first sampling set.
This failure is mainly due to limitations in understanding rare or specialized knowledge, difficulties in handling unusual problem-setting logic, and challenges in context-based reasoning for highly complex exercises.
These exercises have a strong potential to guide model improvement, including targeted knowledge enhancement and refinement of reasoning processes.

Focusing on three commonly used subject categories—STEM, social sciences, and humanities—marked in blue, yellow, and orange in Fig. ~\ref{fig:best_agent}(a), respectively, a total of 6,843 exercises are included in 41 knowledge domains.
The accuracy (Acc) results show that LLMs fail in approximately 1,118 high-difficulty exercises, representing about 16.3\% of the total.

To balance subject diversity and experimental controllability, six representative knowledge domains were selected to form a second sampling subset.
These domains include Machine Learning (n = 20), College Engineering Hydrology (n = 20), Marketing (n = 23), High School Geography (n = 12), Arts (n = 11), and Logic (n = 26).
The selected domains are well distributed across subject categories and difficulty levels.
In addition, the dataset provides clear knowledge-domain labels that support subsequent analysis of agent performance and modeling.

The average accuracy of this secondary sample set is marked with asterisks in the scatter plot in Fig. ~\ref{fig:best_agent}(a).
As shown in the figure, the six representative domains are evenly distributed across the three main subject categories.
These domains fall within the accuracy ranges of [0, 0.3) and [0.3, 0.6), which correspond to ultra-high difficulty and high-difficulty knowledge areas, respectively.
Therefore, they are treated as representative knowledge domains (RKDs) for the three major subject categories.

\subsection{Model Comparisons}
\label{sec:2}
To evaluate the effectiveness and generalization ability of InqEduAgent in learning partner recommendation tasks, a multilevel and multi-strategy experimental framework is designed.
This framework includes several comparison models, ranging from non-modeling support to personalized strategy-based matching.
Seven experimental settings are defined as follows:

\begin{itemize}
    \item \texttt{Baseline}:
    In this setting, agents do not have personality traits.
    There is no subject preference or reasoning style.
    All agents complete tasks independently, without modeling or collaboration.
    This setting serves as a basic reference for evaluating group performance without personalization or coordination.

    \item \texttt{Self-Learning Model (SLM)}:
    Agents are assigned personality traits but complete all tasks independently.
    No interaction is allowed between agents.
    This setting evaluates the direct effect of role modeling on individual learning outcomes.

    \item \texttt{Co-Learning Model (CLM)}:
    Agents are assigned roles and randomly paired with others for collaboration.
    This setting simulates a non-strategic matching process.
    It is used to assess the average benefit of collaboration without intelligent partner selection.

    \item \texttt{InqEduAgent-GP (Global Pareto Matching)}:
    During training, each agent is assigned a predefined role.
    Historical collaboration data are collected across multiple domains.
    A Gaussian Process Regression (GPR) model is trained to capture collaboration gains under different role combinations.
    During testing, the trained GPR model predicts collaboration performance for each pair of candidates based on agent roles and task features.
    A global Pareto front is then constructed using the predicted gains and domain-level performance indicators.
    Communication partners are selected from this front to ensure stable and generalizable matching across domains.

    \item \texttt{InqEduAgent-LP (Localized Pareto Matching)}:
    This variant extends the global model by incorporating each agent’s interaction history.
    A local Pareto front is constructed from the perspective of each agent.
    Partner selection is performed on the basis of this personalized frontier.
    This design improves adaptability to different learning contexts and supports more consistent matching.

    \item \texttt{InqEduAgent-GP (NN)}:
    In this variant, the Gaussian Process Regression module is replaced with a neural network regressor under the global matching framework.
    This modification aims to capture more complex nonlinear collaboration patterns and better model high-dimensional features.

    \item \texttt{InqEduAgent-LP (NN)}:
    This variant integrates neural network regression into the personalized matching framework.
 The history of interaction of each agent is used to construct a local matching mechanism.
    By combining deep modeling with personalized experience, this approach improves adaptability and accuracy in partner selection across tasks.
\end{itemize}

To comprehensively evaluate the modeling capability and generalization performance of InqEduAgent in personalized learning partner recommendation, three evaluation metrics are employed.
These metrics jointly assess accuracy, potential, and stability.
Mean Gain measures the average improvement in learning accuracy across all subjects. It reflects the overall effectiveness and adaptability of the recommendation mechanism. Best Gain represents the upper bound of model performance under optimal matching conditions.
It indicates the potential ceiling and ideal efficiency of the learning framework. Standard Deviation (Std) measures the stability of the model’s recommendations across different partner configurations.
It provides insight into the consistency and robustness of the model in dynamic collaborative learning scenarios.

\begin{table}[t]
\centering
\caption{Model performance across different knowledge domains.}
\label{tab:table1}
\renewcommand{\arraystretch}{1.2}
\setlength{\tabcolsep}{2pt}
\scriptsize
\begin{tabular}{p{0.28\columnwidth} C{0.09\columnwidth} C{0.115\columnwidth} C{0.12\columnwidth} C{0.165\columnwidth} C{0.11\columnwidth}}
\toprule
\textbf{Model} & \textbf{Metric} 
& \textbf{STEM} 
& \textbf{\shortstack{Social\\Science}} 
& \textbf{Humanities} 
& \textbf{Total} \\
\midrule
Baseline         
& Mean & 0.2857 & \textbf{0.2654} & 0.2593 & 0.2704 \\
& Best & 0.4286 & 0.3462 & 0.3704 & 0.3827 \\
& Std  & 0.0607 & 0.0665 & 0.0698 & 0.0452 \\
SLM              
& Mean & 0.3306 & 0.1858 & 0.3111 & 0.2655 \\
& Best & 0.4333 & 0.2558 & 0.4118 & 0.3333 \\
& Std  & 0.0789 & 0.0707 & 0.0671 & 0.0395 \\
CLM              
& Mean & 0.3447 & 0.2139 & 0.3167 & 0.2935 \\
& Best & 0.4314 & 0.3000 & 0.4000 & 0.3548 \\
& Std  & 0.0677 & 0.0412 & 0.0624 & 0.0387 \\
InqEduAgent-GP   
& Mean & 0.3612 & 0.1913 & \textbf{0.3502} & 0.3030 \\
& Best & 0.4286 & 0.3462 & 0.4444 & 0.3704 \\
& Std  & 0.0356 & 0.0648 & 0.0458 & 0.0319 \\
InqEduAgent-LP   
& Mean & \textbf{0.3871} & 0.1896 & 0.3281 & \textbf{0.3047} \\
& Best & 0.4865 & 0.2667 & 0.3784 & 0.3495 \\
& Std  & 0.0546 & 0.0331 & 0.0446 & 0.0325 \\
InqEduAgent-GP (NN) 
& Mean & 0.3438 & 0.2121 & 0.3007 & 0.2873 \\
& Best & 0.4138 & 0.2692 & 0.4074 & 0.3494 \\
& Std  & 0.0626 & 0.0436 & 0.0503 & 0.0297 \\
InqEduAgent-LP (NN) 
& Mean & 0.3611 & 0.1891 & 0.3150 & 0.2910 \\
& Best & 0.4138 & 0.2308 & 0.4444 & 0.3253 \\
& Std  & 0.0357 & 0.0344 & 0.0606 & 0.0214 \\
\bottomrule
\end{tabular}
\end{table}

\begin{table}[t]
\centering
\caption{Ablation Study of Different Model Components}
\label{tab:table2}
\renewcommand{\arraystretch}{1.5}
\setlength{\tabcolsep}{1.5pt}
\scriptsize
\begin{tabular}{@{}p{0.32\columnwidth} C{0.1\columnwidth} C{0.13\columnwidth} C{0.07\columnwidth} C{0.10\columnwidth} C{0.20\columnwidth}@{}}
\toprule
\textbf{Model} 
& \textbf{Role} 
& \textbf{\shortstack{Co-\\learning}} 
& \textbf{GP} 
& \textbf{Pareto} 
& \textbf{\shortstack{Accuracy\\(Gain) \%}} \\
\midrule
Baseline             
& \ding{55} & \ding{55} & \ding{55} & \ding{55} & 26.00 (+0.00) \\
SLM                  
& \ding{51} & \ding{55} & \ding{55} & \ding{55} & 26.55 (+0.55) \\
CLM                  
& \ding{51} & \ding{51} & \ding{55} & \ding{55} & 29.35 (+3.35) \\
InqEduAgent-GP       
& \ding{51} & \ding{51} & \ding{51} & \ding{51} & 30.30 (+4.30) \\
\textbf{InqEduAgent-LP} 
& \ding{51} & \ding{51} & \ding{51} & \ding{51} & \textbf{30.47 (+4.47)} \\
InqEduAgent-GP (NN)   
& \ding{51} & \ding{51} & \ding{55} & \ding{51} & 28.73 (+2.73) \\
InqEduAgent-LP (NN)   
& \ding{51} & \ding{51} & \ding{55} & \ding{51} & 29.10 (+3.10) \\
\bottomrule
\end{tabular}
\end{table}

Table~\ref{tab:table1} reports the performance of seven models across three knowledge domains—STEM, social science, and humanities—as well as their overall results.
Overall, the proposed InqEduAgent variants consistently outperform the baseline and non-strategic collaboration models across most evaluation metrics.
This improvement indicates that the integration of Gaussian process modeling and Pareto frontier optimization enhances the accuracy and adaptability of learning partner recommendations in multi-agent settings.
The results also show that InqEduAgent can capture complex cognitive relationships between learners and exercises, leading to more stable and personalized partner matching.

In terms of Mean Gain, InqEduAgent-LP achieves the highest performance in the STEM domain (0.3871) and in overall results (0.3047).
This result highlights the effectiveness of the localized Pareto matching strategy, which incorporates individual interaction histories to provide more personalized partner recommendations.
By leveraging prior collaboration data, the model dynamically adjusts partner selection to better align with each learner’s cognitive patterns and learning preferences.
This adaptability allows InqEduAgent-LP to perform particularly well in complex tasks that involve high cognitive diversity and interdependent reasoning processes.
Meanwhile, InqEduAgent-GP achieves the highest Mean Gain in the Humanities domain (0.3502).
This result underscores its strong generalization ability in semantically rich and conceptually coherent tasks.
It suggests that the Gaussian process–based mechanism captures meaningful structural relationships between learners and problem spaces.
As a result, the model maintains robust performance even when task semantics vary across domains.
Interestingly, the Baseline model shows a relatively high Mean Gain in the Social Science domain (0.2654).
However, given its weaker performance in other domains, this result is likely caused by random variation in the data distribution rather than true model robustness.
In contrast, both InqEduAgent-LP and InqEduAgent-GP demonstrate consistent improvements across all domains.
These results confirm the advantage of combining global generalization through Gaussian process modeling with local adaptability through Pareto-based optimization.

Regarding Best Gain, InqEduAgent-LP achieves the highest score among all models in the STEM domain (0.4865).
This result demonstrates its strong capability in identifying highly compatible learning partners.
It highlights the strength of personalized and adaptive partner matching, where local Pareto optimization effectively guides the search toward high-quality and mutually beneficial collaborations.
In contrast, InqEduAgent-GP attains the best performance in the Social Science (0.3462) and Humanities (0.4444) domains.
These results further confirm its stability and cross-domain generalization ability.
They suggest that the global matching framework, supported by Gaussian process regression, is effective at capturing shared structural patterns and semantic relationships across diverse learning contexts.
The neural network variants, InqEduAgent-GP (NN) and InqEduAgent-LP (NN), achieve performance close to that of the Gaussian-process-based models in several domains.
For example, GP-NN reaches a Best Gain of 0.4138 in the STEM domain.
These results indicate the potential of neural networks for modeling nonlinear collaboration patterns.
However, under the current dataset size and task complexity, these variants do not surpass InqEduAgent-LP.
This observation suggests that Gaussian processes remain more effective for uncertainty modeling and partner compatibility estimation in small- to medium-scale learning environments.

In terms of recommendation stability, InqEduAgent-GP achieves the lowest overall standard deviation (0.0319).
This result indicates the highest level of robustness and consistency across diverse partner combinations.
It shows that the global Gaussian-process-based matching framework can maintain stable recommendation quality under varying learner profiles and task configurations.
InqEduAgent-LP also demonstrates stable performance across multiple domains.
For example, it achieves a low standard deviation in the Social Science domain (0.0331).
However, slightly higher variability is observed in the STEM domain (0.0546).
This controlled fluctuation reflects the model’s increased sensitivity to individual learning differences.
It allows the system to adjust recommendations based on personal interaction histories.
Notably, the LP-NN variant achieves the lowest overall standard deviation (0.0214).
This result suggests smoother feature representations and more precise internal optimization.
However, its larger fluctuation in the Humanities domain (0.0606) indicates a potential risk of overfitting.
This issue is more likely to occur in semantically complex or data-scarce tasks.

\subsection{Component-removal experiment}
\label{sec:2}
Table~\ref{tab:table2} illustrates the generalizability and flexibility of the proposed model.
The framework consists of four core components: agent roles, communication-based co-learning, Gaussian-process-based decision making, and Pareto-front optimization.
The complete model, InqEduAgent, integrates all four components.
When the Gaussian process and Pareto front are removed, InqEduAgent reduces to the Co-Learning Model (CLM).
Further removing the co-learning module results in the Self-Learning Model (SLM).
Finally, removing agent roles from SLM yields the baseline model.
This step-by-step simplification demonstrates the strong generalization ability of InqEduAgent.
It also reveals the structural consistency of the model design through progressive module removal.

Each module in InqEduAgent is replaceable, which demonstrates the flexibility of the framework.
When the Gaussian process is replaced with a neural network (NN), the model becomes InqEduAgent (NN).
The results show that all InqEduAgent variants outperform their NN-based counterparts.
This finding confirms that Gaussian processes are more effective for uncertainty modeling and predictive reasoning in this task.
Similarly, replacing the local Pareto front with the global Pareto front transforms InqEduAgent-LP into InqEduAgent-GP.
InqEduAgent-LP achieves an accuracy that is 0.17 higher than that of InqEduAgent-GP.
Although the improvement is moderate, it is stable across settings and highlights the robustness of the proposed framework.

\section{Conclusion}
This paper proposes InqEduAgent for inquiry-oriented learning. The model supports both human–AI collaboration and human–human collaboration. In InqEduAgent, non-parametric modeling enhanced by Gaussian processes is used to explore interaction patterns among learners and between learners and domain knowledge. These patterns are learned from historical interaction data and are used to build a prediction model. Based on this model, InqEduAgent can perform adaptive learning-partner matching for newly presented questions, even when explicit domain labels are unavailable. The system identifies communication partners who are most helpful for improving a learner’s understanding of the target knowledge. Results from comparative experiments show that InqEduAgent achieves good performance across most knowledge domains. The findings of this study can be applied to optimize intelligent matching systems in human–AI co-learning platforms.

\textbf{Data Availability Statement:} All code, datasets, and appendices are publicly available at \url{https://github.com/InqEduAgent/InqEduAgent}.

\bibliographystyle{IEEEtran}
\bibliography{reference}

\end{document}